\documentclass{article}
\usepackage[preprint]{neurips_2025}
\usepackage[utf8]{inputenc} %
\usepackage[T1]{fontenc}    %
\usepackage{times,latexsym}
\usepackage{graphicx} \graphicspath{{figures/}}
\usepackage{amsmath,amssymb,mathabx,mathtools,amsthm,nicefrac}
\usepackage[linesnumbered,ruled,vlined]{algorithm2e}
\usepackage{acronym}
\usepackage{enumitem}
\usepackage[pagebackref,breaklinks,colorlinks]{hyperref}
\usepackage{balance}
\usepackage{xspace}
\usepackage{wrapfig}
\usepackage{setspace}
\usepackage[skip=3pt,font=small]{subcaption}
\usepackage[skip=3pt,font=small]{caption}
\usepackage[dvipsnames,svgnames,x11names,table]{xcolor}
\usepackage[capitalise,noabbrev,nameinlink]{cleveref}
\usepackage{booktabs,tabularx,colortbl,multirow,multicol,array,makecell,tabularray}
\usepackage{overpic,wrapfig}
\usepackage[misc]{ifsym}
\usepackage{pifont}
\usepackage{tikz}
\usepackage{siunitx}

\makeatletter
\DeclareRobustCommand\onedot{\futurelet\@let@token\@onedot}
\def\@onedot{\ifx\@let@token.\else.\null\fi\xspace}
\def\eg{\emph{e.g}\onedot}

\def\etal{\emph{et al}\onedot}
\makeatother

\DeclareMathOperator*{\argmin}{arg\,min}

\makeatletter
\renewcommand{\paragraph}{%
  \@startsection{paragraph}{4}%
  {\z@}{0ex \@plus 0ex \@minus 0ex}{-1em}%
  {\hskip\parindent\normalfont\normalsize\bfseries}%
}
\makeatother

\crefname{algorithm}{Alg.}{Algs.}
\Crefname{algocf}{Algorithm}{Algorithms}
\crefname{section}{Sec.}{Secs.}
\Crefname{section}{Section}{Sections}
\crefname{table}{Tab.}{Tabs.}
\Crefname{table}{Table}{Tables}
\crefname{figure}{Fig.}{Figs.}
\Crefname{figure}{Figure}{Figures}
\crefname{equation}{Eq.}{Eqs.}
\Crefname{equation}{Equation}{Equations}
\crefname{appendix}{Appx.}{Appxs.}
\Crefname{appendix}{Appendix}{Appendices}

\definecolor{gblue}{HTML}{4285F4}
\definecolor{gred}{HTML}{DB4437}
\definecolor{ggreen}{HTML}{0F9D58}

\acrodef{bigai}[BIGAI]{Beijing Institute for General Artificial Intelligence}
\acrodef{pku}[PKU]{Peking University}
\acrodef{thu}[THU]{Tsinghua University}
\acrodef{ddpm}[DDPM]{Denoising Diffusion Probabilistic Model}
\acrodef{mala}[MALA]{Metropolis-Adjusted Langevin Algorithm}
\acrodef{sdf}[SDF]{Signed Distance Function}
\acrodef{fkine}[FK]{Forward Kinematics}
\acrodef{ik}[IK]{Inverse Kinematics}
\acrodef{ppo}[PPO]{Proximal Policy Optimization}
\acrodef{rl}[RL]{Reinforcement Learning}
\acrodef{gcrl}[GCRL]{Goal-Conditioned Reinforcement Learning}
\acrodef{mdp}[MDP]{Markov Decision Process}
\acrodef{ood}[OOD]{Out-of-Domain}
\acrodef{ibs}[IBS]{Intersection Bisector Surface}
\acrodef{bp}[BP]{Behavior-Paradigm}
\acrodef{dfc}[DFC]{Differentiable Force Closure}
\acrodef{gws}[GWS]{Grasp Wrench Space}
\acrodef{attr-diffusion}[AD]{Attraction-Diffusion}
\acrodef{adtd}[ADTD]{Attraction-Diffusion Transition Discovery}
\acrodef{mcmc}[MCMC]{Markov-Chain Monte-Carlo}
\acrodef{ipc}[IPC]{Incremental Potential Contact}
\acrodef{ip}[IP]{Incremental Potential}
\acrodef{abd}[ABD]{Affine Body Dynamics}
\acrodef{mpm}[MPM]{Material Point Methods}
\acrodef{fem}[FEM]{Finite Element Methods}
\acrodef{pbd}[PBD]{Position-based Dynamics}
\acrodef{vbts}[VBTS]{vision-based tactile sensor}
\acrodef{urdf}[URDF]{Unified Robot Description Format}
\acrodef{ccd}[CCD]{Continuous Collision Detection}
\acrodef{sdg}[SDG]{Synthetic Data Generation}
\acrodef{dof}[DoF]{Degree of Freedom}
\acrodef{dnn}[DNN]{Deep Neural Network}

\newcommand{\simName}{\texttt{\textbf{Taccel}}\xspace}

\usepackage{bm}

\newcommand\bx{\mathbf{x}}
\newcommand\bs{\mathbf{s}}
\newcommand\bS{\mathbf{S}}
\newcommand\dbx{\dot{\mathbf{x}}}

\newcommand\by{\mathbf{y}}
\newcommand\dby{\dot{\mathbf{y}}}

\newcommand\bM{\mathbf{M}}
\newcommand\bP{\mathbf{P}}
\newcommand\bJ{\mathbf{J}}

\newcommand\calE{\mathcal{E}}
\newcommand\calS{\mathcal{S}}
\newcommand\calG{\mathcal{G}}

\newcommand\calQ{\mathcal{Q}}

\newcommand\Dt{\Delta t}

\newcommand{\vf}[1]{{\bm{#1}}}
\newcommand{\mf}[1]{{\mathbf{#1}}}

\title{\simName: Scaling Up Vision-based Tactile Robotics via High-performance GPU Simulation}

\author{%
    \normalfont\setlength{\tabcolsep}{3pt}%
    \hspace{-1.5em}\begin{tabular}{ccc}
        \textbf{Yuyang Li}$^{\,1,2,3,4,\,\star}$ & \textbf{Wenxin Du}$^{\,5,\,\star}$ & \textbf{Chang Yu}$^{\,5,\,\star}$  \\
        \texttt{\footnotesize{}y.li@stu.pku.edu.cn} & \texttt{\footnotesize{}setsunainn@gmail.com} & \texttt{\footnotesize{}changyu1@g.ucla.edu}
    \end{tabular}
    \vspace{6pt}\\
    \hspace{-1.5em}\begin{tabular}{ccc}
        \textbf{Puhao Li}$^{\,4}$ & \textbf{Zihang Zhao}$^{\,1,2,3}$ & \textbf{Tengyu Liu}$^{\,4}$ \\
        \texttt{\footnotesize{}liph23@mails.tsinghua.edu.cn} &  \texttt{\footnotesize{}zhaozihang@stu.pku.edu.cn} & \texttt{\footnotesize{}liutengyu@bigai.ai}
    \end{tabular}
    \vspace{6pt}\\
    \hspace{-1.5em}\begin{tabular}{ccc}
        \textbf{Chenfanfu Jiang}$^{\,5,\,\textrm{\Letter}}$ & \textbf{Yixin Zhu$^{\,1,2,3,6,\,\textrm{\Letter}}$} & \textbf{Siyuan Huang$^{\,4,\,\textrm{\Letter}}$}  \\
        \texttt{\footnotesize{}cffjiang@math.ucla.edu} &  \texttt{\footnotesize{}yixin.zhu@pku.edu.cn} &  \texttt{\footnotesize{}huangsiyuan@ucla.edu}
    \end{tabular}
    \vspace{6pt}\\
    $^\star$Equal contributor \quad $^{\textrm{\Letter}}$Corresponding author\\
    \footnotesize $^1$ Institute for Artificial Intelligence, Peking University\\
    \footnotesize $^2$ School of Psychological and Cognitive Sciences, Peking University\\
    \footnotesize $^3$ Beijing Key Laboratory of Behavior and Mental Health, Peking University\\
    \footnotesize $^4$ State Key Lab of General AI, Beijing Institute for General AI\\
    \footnotesize $^5$ AIVC Laboratory, University of California, Los Angeles\\
    \footnotesize $^6$ Embodied Intelligence Lab, PKU-Wuhan Institute for Artificial Intelligence
    \vspace{3pt}\\
    \url{https://taccel-simulator.github.io}
    \vspace{-18pt}
}

\begin{document}

\maketitle

\begin{abstract}
Tactile sensing is crucial for achieving human-level robotic capabilities in manipulation tasks~\cite{westling1984factors}.
As a promising solution, \Acfp{vbts}~\cite{yuan2017gelsight,lin20239dtact} offer high spatial resolution and cost-effectiveness, but present unique challenges in robotics for their complex physical characteristics and visual signal processing requirements.
The lack of efficient and accurate simulation tools for \acp{vbts} has significantly limited the scale and scope of tactile robotics research~\cite{wang2022tacto,chen2024general}.
We present \simName, a high-performance simulation platform that integrates \ac{ipc} and \ac{abd} to model robots, tactile sensors, and objects with both accuracy and unprecedented speed, achieving an 18-fold acceleration over real-time across thousands of parallel environments.
Unlike previous simulators that operate at sub-real-time speeds with limited parallelization, \simName provides precise physics simulation and realistic tactile signals while supporting flexible robot-sensor configurations through user-friendly APIs. Through extensive validation in object recognition, robotic grasping, and articulated object manipulation, we demonstrate precise simulation and successful sim-to-real transfer.
These capabilities position \simName as a powerful tool for scaling up tactile robotics research and development, potentially transforming how robots interact with and understand their physical environment.
\end{abstract}

\section{Introduction}

The ability to physically interact with the environment through touch is fundamental to robotic manipulation~\cite{billard2019trends,edmonds2019tale}. While vision provides global scene understanding, tactile sensing captures crucial local contact information~\cite{xu2023visual} essential for precise manipulation. Among various tactile sensing technologies~\cite{zhanat2015tactile,jalgha2012hierarchical,liu2012tactile,jara2014control}, \acfp{vbts} such as GelSight~\cite{yuan2017gelsight} and 9DTact~\cite{lin20239dtact} have emerged as a central focus in tactile research. Their ability to provide high-resolution tactile feedback through camera-captured deformation patterns of elastic gel pads, combined with cost-effectiveness, has driven significant advances in robotics~\cite{zhao2024tac,li2024minitac,si2024difftactile,chen2024general,zhao2024embedding}.

The primary challenge in scaling up \ac{vbts}-equipped robot simulation lies in accurately modeling the hyperelastic soft gel pad and its contact~\cite{zhao2024tac,du2024tacipc}. Current approaches follow two main directions: rigid-body approximations~\cite{wang2022tacto,xu2023efficient} and soft-body simulations~\cite{si2024difftactile,chen2024general,du2024tacipc,hogan2020tactile,kim2024texterity,zhao2024tac}. While rigid-body methods efficiently support basic tasks like pick-and-place~\cite{akinola2024tacsl,wang2022tacto}, they cannot capture the fine-grained contactsd and elastomer deformations essential for complex manipulation tasks~\cite{zhao2024tac,du2024tacipc} and detailed force distribution analysis~\cite{lloyd2024pose,si2024difftactile}. Soft-body simulations offer higher fidelity but face significant computational challenges that limit their practical application in large-scale experiments.

\begin{wrapfigure}{r}{0.38\textwidth}
  \centering
  \includegraphics[width=\linewidth]{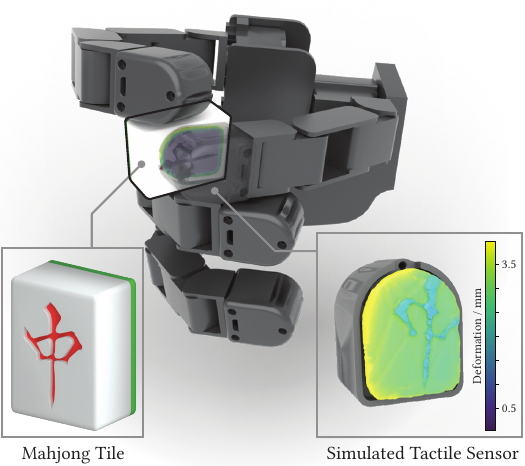}
  \caption{\textbf{\simName demonstration of tactile robotics simulation.} An Allegro Hand with four \acsp{vbts} performing a precision grasp on a mahjong tile. The deformation map precisely captures the tile's surface geometries.}
  \label{fig:teaser}
  \vspace{-20pt}
\end{wrapfigure}

An ideal \ac{vbts} simulator must simultaneously achieve:
\begin{itemize}[leftmargin=*,nolistsep,noitemsep]
    \item \textbf{Precision}: Precise modeling of robots, sensors, and objects with physically valid solutions, particularly maintaining inversion-free and intersection-free states during complex contact interactions; generation of realistic tactile signals across multiple resolutions, from high-resolution RGB patterns and depth maps to low-resolution marker movements.
    \item \textbf{Scalability}: Capability for large-scale parallelization for extensive simulation data generation.
    \item \textbf{Flexibility}: Support for diverse robotic platforms and sensor configurations, from parallel grippers to multi-finger hands with varying sensor arrangements.
\end{itemize}

\begin{table*}[b!]
    \centering
    \small
    \caption{\textbf{Comprehensive comparison of \acs{fem}-based \acs{vbts} simulators.} \textbf{Soft Mat.}: modeling of deformable materials (\acs{fem}: \Acl{fem}). \textbf{Stiff Mat.}: modeling of stiff materials (Rigid: rigid body, \acs{abd}: \Acl{abd}, \acs{mpm}: \Acl{mpm}, \acs{pbd}: \Acl{pbd}). Contact: collision handling method (Virtual: approximated contact, Penalty: penalty-based, \ac{ipc}: Incremental Potential Contact). \textbf{RGB Signal}: RGB tactile pattern generation method (Look-up: look-up tables, \acs{dnn}: \Acl{dnn}, \ding{55}: not supported). \textbf{Robot}: range of supported robotic systems. The last two columns report parallel simulation capabilities: maximum parallel environments and simulation speed relative to real-time in a peg insertion test, with dual sensors in low/high resolutions, measured on an NVIDIA H100 80G GPU. Details are in \cref{fig:perf}.}
    \resizebox{\linewidth}{!}{
    \begin{tabular}{cccccccc}
        \toprule
        \textbf{Simulator} & \textbf{Soft Mat.} & \textbf{Stiff Mat.} & \textbf{Contact} & \textbf{RGB Signal} & \textbf{Robot} & \textbf{\#~Env}\,$\uparrow$ & \textbf{Sim Speed}\,$\uparrow$ \\
        \midrule
        Taxim~\cite{si2022taxim} & - & Rigid & Virtual & Look-up & Sensor & 1 & - \\ 
        DiffTactile~\cite{si2024difftactile} & \acs{fem} & \acs{mpm}/\acs{pbd} & Penalty & \acs{dnn} & Gripper & 1 & - \\ 
        SAPIEN-\acs{ipc}~\cite{chen2024general} & \acs{fem} & \textbf{\acs{abd}} & \textbf{\acs{ipc}} & \ding{55} & Gripper & 256 / \textit{4} & 0.81$\times$ / 0.03$\times$ \\ 
        \midrule
        \textbf{\simName (Ours)} & \acs{fem} & \textbf{\acs{abd}} & \textbf{\acs{ipc}} & \textbf{\acs{dnn}} & \textbf{Any} & \textbf{\textit{4096} / \textit{64}} & \textbf{\textit{18.30$\times$} / 0.25$\times$} \\ 
        \bottomrule
        \label{tab:vbts-sim-summary}
    \end{tabular}
    }
\end{table*}

As detailed in \cref{tab:vbts-sim-summary}, existing solutions often compromise on precision, scalability, or flexibility. They typically produce suboptimal physics, operate slower than real-time with limited parallel environments, or focus on specific sensor setups or simple grippers. These limitations significantly impede the broader application of tactile robotics.

To address these challenges, we present \simName, a high-performance simulation platform for scaling up robots with \ac{vbts}-integration. Built on state-of-the-art simulation techniques, \simName provides dedicated components for simulating robots (\cref{sec:vtbs_sim}), tactile sensors (\cref{sec:tactile-sim}), and tactile signal generation (\cref{sec:apis}). Comprehensive evaluations (\cref{sec:sim:benchmark}) demonstrate its characteristics:
\begin{itemize}[leftmargin=*,nolistsep,noitemsep]
    \item \textbf{Precision}: \simName leverages advanced solid material simulation techniques (\ac{ipc}~\cite{li2020incremental} and \ac{abd}~\cite{lan2022affine}) for physical accuracy. \ac{ipc} guarantees inversion- and intersection-free contact solutions, while the integration of \ac{abd} allows for efficient and precise simulation.
    \item \textbf{Scalability}: With an efficient \ac{abd}-\ac{ipc} implementation with NVIDIA warp~\cite{warp2022}, \simName achieves unprecedented parallelization. On a single H100 GPU, it reaches over 900 FPS (4096 environments, 18\(\times\) wallclock time) for a peg-insertion task with dual sensors and 12.67 FPS (256 environments, 0.25\(\times\) wallclock time) in a dexterous grasping task with full-hand tactile sensing.
    \item \textbf{Flexibility}: \simName provides user-friendly APIs for seamless integration of diverse robotic platforms and sensor configurations. Users can easily load and configure robots through \ac{urdf} with auxiliary configurations, supporting applications from simple grippers to complex manipulation tasks, like the mahjong tile sensing task in \cref{fig:teaser}.
\end{itemize}

\simName's utility is validated through three fundamental tactile-informed robotic tasks (\cref{sec:robotics}). In object classification, models trained solely on \simName's synthetic tactile signals demonstrate strong generalization to real-world data without adaptation. In grasping experiments across four robotic hand designs, we showcase the platform's versatility in handling diverse robot configurations and tactile signal types. In articulated object manipulation tasks, we demonstrate \simName's physical fidelity through close correspondence between simulated and real-world robot behavior.

Our key contributions include: (i) development of a high-performance simulation platform combining precise physics, realistic tactile signal generation, and massive parallelization; (ii) user-friendly APIs enabling flexible robot-sensor integration and high-fidelity tactile signal synthesis; (iii) comprehensive evaluation of the platform's precision and scalability; and (iv) extensive experimental validation across diverse tactile robotic tasks. By enabling large-scale, high-fidelity simulation of \ac{vbts}-equipped robots, \simName aims to accelerate future research in tactile robotics.

\section{Related Work}

Early \ac{vbts} simulators focused on normal deformation scenarios, approximating hyperelastic behavior through geometric computations and surface modifications~\cite{gomes2019gelsight,wang2022tacto,xu2023efficient,akinola2024tacsl,si2022taxim}. While efficient in generating high-resolution tactile signals through physics-based rendering or look-up tables, these approaches inadequately capture elastomer dynamics during complex manipulation tasks, especially those involving tangential forces and continuous interactions~\cite{zhao2024tac}.

Recent approaches have achieved higher physical fidelity by incorporating advanced solid material simulation techniques. Methods using \ac{mpm}~\cite{sulsky1994particle,hu2018moving,hu2019chainqueen} and \ac{fem}~\cite{li2020incremental,lan2022affine} better model elastomer properties through time-integrated deformation computations~\cite{chen2023tacchi,chen2024general,si2024difftactile,du2024tacipc}. Notable improvements include the adoption of \ac{ipc}~\cite{li2020incremental} by several simulators~\cite{du2024tacipc,chen2024general}, providing robust contact handling with guaranteed inversion- and intersection-free solutions. \cref{tab:vbts-sim-summary} compares key features of representative approaches.

\simName builds on these advances by combining \ac{ipc} and \ac{abd} in a unified platform, achieving both physical accuracy and computational efficiency while supporting diverse robot configurations and enabling large-scale parallel simulation for robot learning applications.

A more comprehensive review on related works is in \cref{supp:related} concerning the \acp{vbts} and tactile-informed robotic tasks.

\section{Unified \texorpdfstring{\acs{ipc}}{} Simulation in \simName}\label{sec:ipc}

This section presents the unified \ac{ipc} simulation framework in \simName, detailing its mathematical foundations. For complete derivations, we refer readers to \cref{sec:ipc} and the original works~\cite{lan2022affine,li2020incremental,chen2022unified}.

\subsection{Problem Formulation and Soft Body Dynamics}

We consider $n_s$ tetrahedralized soft bodies discretized into $N_s$ vertices with positions $\vf{x}_1, \vf{x}_2, ..., \vf{x}_{N_s}$ in Cartesian space. The system state is represented by the stacked position vector $\bx = [\vf{x}^T_1, \vf{x}^T_2, ..., \vf{x}^T_{N_s}]^T \in \mathbb{R}^{3N_s}$. Following Lagrangian mechanics, we express the system's Lagrangian as $L(\bx, \dot{\bx}) = T(\bx, \dot{\bx}) - V(\bx)$, where $T(\bx, \dot{\bx}) = \frac{1}{2} \dot{\bx}^T \bM \dot{\bx}$ represents kinetic energy with mass matrix $\bM \in \mathbb{R}^{3N_s\times3N_s}$. The potential energy $V(\bx)$ comprises two terms: an elastic energy $\Phi(\bx)$ utilizing the Neo-Hookean constitutive model for hyperelastic materials (characterized by Young's modulus $\calE$ and Poisson's ratio $\nu$), and external forces $E_{\text{ext}}(\bx)$.

\subsection{Frictional Contact}\label{sec:sim:time-stepping}
The Euler–Lagrange equation $\frac{\partial L}{\partial \bx}(\bx, \dot{\bx}) - \frac{d}{dt}\frac{\partial L}{\partial \dot{\bx}}(\bx, \dot{\bx}) = 0$ is equivalent to the following Incremental Potential (IP) energy minimization problem~\cite{li2020incremental} under the backward Euler integration scheme: 
\begin{equation}
    \begin{aligned}
        E_{\text{IP}}(\bx) &= \frac{1}{2}(\bx - \bx^{n} - \Dt\dot{\bx}^{n})^T\bM(\bx - \bx^{n} - \Dt\dot{\bx}^{n})
        + \Dt^2V(\bx), \\ 
        \bx^{n+1} &= \argmin_{\bx} E_{\text{IP}}(\bx),
    \end{aligned}
\end{equation}
where time is discretized into steps $\{t_n = n\Dt: n\in \mathbb{N}\}$ with a fixed step size $\Dt > 0$, and $\bx^{n}=\bx(t_n)$. 

To ensure intersection-free trajectories, IPC augments the objective with log barrier functions $b(d_k(\bx))$, which diverge to $+\infty$ as the distance $d_k(\bx)$ between contact primitive pair $k$ approaches zero. Additionally, IPC introduces an approximate frictional potential energy $D_k(\bx, \bx^{n})$ that captures frictional forces through its gradient. The full simulation thus minimizes the IPC energy: 
\begin{equation}
\begin{aligned}
    E_{\text{\ac{ipc}}}(\bx) &= E_{\text{IP}}(\bx) + \Dt^2B(\bx) + \Dt^2D(\bx, \bx^{n}). 
    \label{eqn:fullspace_ipc}
\end{aligned}
\end{equation}
with {\small$B(\bx)=\kappa \sum_{k\in \mathcal{B}} A_k b(d_k(\bx))$, $D(\bx, \bx^n)=\sum_{k\in\mathcal{B}} D_k(\bx, \bx^n)$}, where $\kappa > 0$ controls contact stiffness. 

\subsection{\texorpdfstring{\acs{abd}}{} and Unified Simulation}\label{sec:affine}

For $n_a$ affine bodies, we introduce a reduced coordinate space $\by \in \mathbb{R}^{12 n_a}$ with an embedding map $\phi: \mathbb{R}^{12 n_a} \rightarrow \mathbb{R}^{3N_a}$ that projects reduced coordinates to full-space vertices $\phi(\mf{\by})$~\cite{lan2022affine}, where $N_a$ denotes the total vertex count of affine bodies' surface meshes. Each affine body uses 12 \ac{dof}: three for translation ($\mathbb{R}^{3}$) and nine for affine deformation ($\mathbb{R}^{3\times 3}$).
\begin{equation}
    \begin{aligned}
        T(\by, \dot{\by}) = \frac{1}{2} \dot{\bx}^T \bM \dot{\bx} = \frac{1}{2} \dot{\phi}(\by)^T \bM\dot{\phi}(\by) = \frac{1}{2} (\bJ\dot{\by})^T \bM (\bJ\dot{\by}) = \frac{1}{2}\dot{\by}^T(\bJ^T\bM\bJ)\dot{\by}= \frac{1}{2}\dot{\by}^T \mf{M^y} \dot{\by},
    \end{aligned}
\end{equation}
where $\bJ = \frac{\partial \phi}{\partial \by}\in \mathbb{R}^{3N_a \times 12 n_a }$ is the Jacobian, $\bM$ is the full-space mass matrix, and $\mf{M^y} = \bJ^T\bM\bJ$ is the reduced-space mass matrix. The potential energy $V(\by)$ includes an As-Rigid-As-Possible (ARAP) term $\Phi^{\by}(\by)=\Phi^{\bx}(\phi(\bx))$ with large $\kappa_s$ to limit deformation, plus external forces $E_{\text{ext}}(\by)$.

Combining with \cref{eqn:fullspace_ipc} yields the unified \ac{ipc} energy~\cite{chen2022unified} for the full system state $\{\by; \bx \}\in \mathbb{R}^{12 n_a + 3 N_s}$:
\begin{equation}
    \begin{aligned}
        E_{\text{\ac{ipc}}}(\by; \bx) &= E_{\text{IP}}(\bx) + E_{\text{IP}}(\by)+ \Dt^2B(\phi(\by); \bx) + \Dt^2D(\phi(\by); \bx, \phi(\by^{n}); \bx^{n}), \\
        E_{\text{IP}}(\by) &= \frac{1}{2}(\by - \by^{n} - \Dt\dot{\by}^{n})^T\mf{M^y}(\by - \by^{n} - \Dt\dot{\by}^{n}) + \Dt^2V(\by).
    \end{aligned}
    \label{eqn:unified_ipc}
\end{equation}

The next timestep's configuration follows from minimizing the following barrier-augmented IP:
\begin{equation}
    \by^{n+1};\bx^{n+1} = \argmin_{\by;\bx} E_{\text{\ac{ipc}}}(\by;\bx).
    \label{eqn:unified_ipc_variational}
\end{equation}

\subsection{Kinematic Constraints}\label{sec:sim:kine-constraint}

We express kinematic constraints as $\bS^{\bx} \bx = \bs^{\bx}$ and $\bS^{\by} \by = \bs^{\by}$, where $\bS^{\bx} \in \mathbb{R}^{c^{\bx} \times 3 N_s}, \bs^{\bx} \in \mathbb{R}^{3 N_s}$ for soft bodies, and $\bS^{\by} \in \mathbb{R}^{c^{\by} \times 12 n_a}, \bs^{\by} \in \mathbb{R}^{12 n_a}$ for affine bodies. To enforce the constraints, we augment $E_{\text{\ac{ipc}}}$ using the Augmented Lagrangian method with Lagrangian multipliers $\mf{\lambda}^{\bx} \in \mathbb{R}^{c^{\bx}}, \mf{\lambda}^{\by} \in \mathbb{R}^{c^{\by}}$:
\begin{equation}
        E_{\text{\ac{ipc}}}^{\text{AL}}(\by; \bx) = E_{\text{\ac{ipc}}}(\by; \bx) +  \Vert (\bS^{\bx} \bx - \bs^{\bx} )^T\mf{\lambda}^{\bx}\Vert_2^2 + \Vert (\bS^{\by} \by - \bs^{\by} )^T\mf{\lambda}^{\by}\Vert_2^2.
    \label{eqn:unified_ipc_AL}
\end{equation}

\section{Robot and \texorpdfstring{\acs{vbts}}{} Simulation in \simName}\label{sec:vtbs_sim}

Building upon the unified ABD-IPC, \simName implements robot and \ac{vbts} simulation through a modular design that leverages the complementary strengths of affine- and soft-body dynamics.

\subsection{Robot and Sensor Simulation}

\paragraph*{Robot Modeling}

For a robot with $D$-DoF, $L$ links, and $N$ \acfp{vbts}, \simName constructs its kinematic model from a \ac{urdf} specification, loading visual and collision meshes as affine bodies (\cref{sec:affine}). Given the robot's global transformation $T_r$ and joint configuration $q \in \mathcal{Q}$, forward kinematics yields the transformation of each link $j$ in the world frame, $\prescript{r}{l_j}{T}(q)$.

\paragraph*{Tactile Sensors Modeling}

Each \ac{vbts} is simulated by a tetrahedral mesh as a soft volumetric body (\cref{sec:ipc}) for its gel pad, attached to its corresponding robot link. For the $i$-th sensor's gel pad $\calG_i$ attached to link $l_j$ with local transformation $\prescript{l_j}{\calG_i}{T}$, we denote its outer surface as $\calS_i = \partial \calG$. The surface comprises a reflective-coated region $\partial^+ \calG_i$ and a sensor-attached region $\partial^- \calG$. Contact interactions during simulation cause gel pad deformation, transforming the coated surface $\partial^+ \calG$ to $\partial^+ \tilde{\calG}$ and marker positions to $\tilde{\bP}_i$. These deformed quantities serve as the foundation for generating multiple types of tactile signals, each suited for different robotic applications.

\paragraph*{Robot Actions}

For scene initialization, we compute affine states through forward kinematics for robot links and explicit state specification for stiff objects. Gel pad node positions are transformed to the world frame, with all states written to $\bx, \by$, and velocities $\dbx, \dby$ initialized to zero. Robot actions are implemented through kinematic constraints. From joint space targets, we compute affine state targets for links and node position targets for gel pad attached surfaces $\partial^- \calG$, assembled as $\bs^\bx, \bs^\by$. Selection matrices $\bS^\bx, \bS^\by$ apply these constraints, with remaining states solved via time stepping.

\subsection{Tactile Signal Simulation in \simName}\label{sec:tactile-sim}

While many works directly generate tactile signals from the object geometry and frictional forces during rigid body simulation for efficiency~\cite{wang2022tacto,xu2023efficient,akinola2024tacsl}, this approach can lead to inauthentic signals in dynamic scenarios. Instead, \simName supports accurate simulation of high-resolution soft bodies to fully capture the fine-grained contact and deformation patterns.

\paragraph{RGB Images and Depth Maps}

High-resolution tactile signals, including RGB images and depth maps, are essential when fine-grained details like object texture and local geometries are required. The signal generation process proceeds in two stages. First, we extract the depth and normal maps $d_{(u,v)}, n_{(u,v)}$ for pixel coordinates $(u,v)$ from the deformed coated surface $\partial^+ \tilde{\calG}$. Next, following the method of Si \etal~\cite{si2024difftactile}, we apply a \ac{dnn} to generate RGB tactile signals from the depth information. Specifically, for each pixel coordinate $(u,v)$, a pixel-to-pixel \ac{dnn} parameterized by $\theta$ maps the inputs to the pixel colors relative to a reference image (RGB signal of undeformation gel pad): $f_\theta \left(\gamma(u, v), n_{(u, v)} \right) \mapsto \Delta \sigma_{(u, v)}$, which are added to the reference image to obtain the final RGB image. $\gamma(\cdot, \cdot)$ provides the 2D positional encoding of the pixel coordinate. The model is trained on patches from 200 real tactile images and corresponding depth map annotations.

\paragraph{Markers} Within $\partial^+ \calG_i$, $m_i$ markers are positioned at locations $\prescript{\calG_i}{}{\bP}_i = \{ p^{(i)}_k \in \mathbb{R}^3, k = 1, \dots, m_i \}_{i=1}^N$, each defined by barycentric coordinates in its triangle: $p^{(i)}_k = \sum_{u=1}^3 \alpha_u\bx^{(i,k)}_u$, where $\sum_{u=1}^3 \alpha_u = 1, \alpha_u \in [0, 1]$. Low-resolution tactile signals primarily track marker positions and their movements by computing their new positions throughout the simulation (at time step $t$): $\tilde{p}^{(i)}_k (t) = \sum_{u=1}^3 \alpha_u \tilde{\bx}^{(i,k)}_u (t)$,
and projecting them on the tactile image. The marker flows, representing local deformation patterns, are then computed as: $\Delta \bP_i = \{ \tilde{p}^{(i)}_k - p^{(i)}_k \} = \{ \Delta p_k^{(i)} \}$.

\paragraph{3D Tactile Signals} The depth map and marker positions can be transformed into a dense or sparse 3D point cloud in the world frame using robot kinematics and sensor configurations. These 3D tactile signals provide crucial spatial information for robotic manipulation tasks. We demonstrate their effectiveness through the simulation of Tac-Man framework~\cite{zhao2024tac} in \cref{sec:rob:manip}.

\begin{figure*}[t!]
    \centering
    \includegraphics[width=\linewidth]{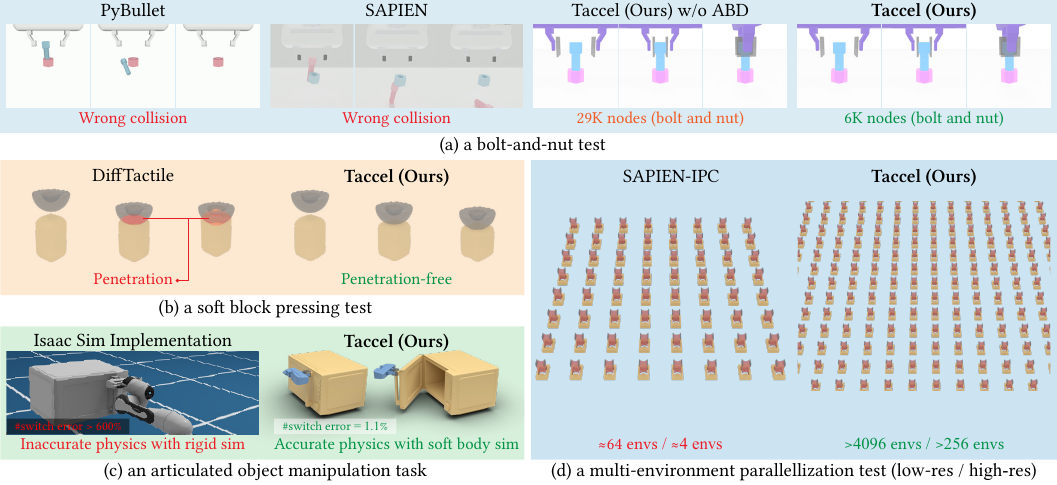}
    \caption{\textbf{Comprehensive evaluation of physics simulation capabilities across \ac{vbts} simulators.} (a) Bolt-nut assembly involving contact between non-convex objects, where \simName achieves stable simulation; (b) Soft block pressing test with soft-soft contacts, where \simName maintains penetration-free interactions; (c) Tactile-informed articulated object manipulation, where \simName replicates real-world interactions with \~1\% physical error; (d) Parallel environment test using a peg insertion task, showing \simName's scalability.}
    \label{fig:sim-compare}
\end{figure*}

 We further demonstrate \simName's capability to scale up synthetic data generation of tactile signals via robotic grasping simulations and explore object recognition model learning (\cref{sec:rob:cognition,sec:rob:grasp}).

\subsection{API Designs in \simName}\label{sec:apis}

\simName provides intuitive Python APIs designed to make tactile robotics simulation accessible to researchers while maintaining high performance through NVIDIA Warp~\cite{warp2022}. The APIs allows for seamless loading of robots from \ac{urdf} files with automatic parsing, sensor configurations from auxiliary files, and objects from mesh files. Users can efficiently reset simulation states or apply kinematic targets to control the robots in familiar formats (NumPy arrays, PyTorch tensors). Further, \simName supports parallel simulation of multiple environments similar to IsaacGym~\cite{makoviychuk2021isaac}.

To foster community development, we will release \simName codebase and documentations, while maintaining active collaboration with researchers to incorporate feedback, add features, and expand capabilities, ensuring its evolution as a comprehensive tool for tactile robotics research.

\section{Performance Evaluation of \simName}\label{sec:sim:benchmark}

We evaluate \simName through comprehensive benchmarks on its precision and efficiency. Our analysis demonstrates that the combination of \ac{abd} and \ac{ipc} provides significant advantages over existing approaches (\cref{sec:eval:simulation}), generates high-quality tactile signals (\cref{sec:eval:tactile-signal}), ensures precise frictions and deformation solving for the gel pad (\cref{sec:eval:friction}), and enables efficient scaling (\cref{sec:eval:scaling}).

\subsection{Overall Comparisons}\label{sec:eval:simulation}

As illustrated in \cref{fig:sim-compare}, \simName achieves superior precision and efficiency through its unified \ac{abd} and \ac{ipc} framework, demonstrated across four challenging scenarios.

First, the bolt screwing task (\cref{fig:sim-compare}a) demonstrates \simName's ability to handle complex physical interactions between highly non-convex objects, where conventional simulators including PyBullet and SAPIEN~\cite{xiang2020sapien} fail to handle. This capability stems from our \ac{abd} formulation, which provides an efficient approach to simulating dense rigid-soft body interactions. Alternative approaches either model stiff objects as soft bodies, introducing excessive \acp{dof} and computational overhead (\cref{fig:sim-compare}a, ours w/o ABD), or rely on RBD~\cite{ferguson2021intersection}, which requires expensive nonlinear \ac{ccd} calculations to avoid intersection, significantly degrading performance.

Next, physical realism in \simName is demonstrated through the soft block pressing scenario (\cref{fig:sim-compare}b), where our collision-free and intersection-free guarantees produce notably more realistic deformations compared to penalty-based approaches. This precision extends to practical robotics applications, as shown in the Tac-Man microwave manipulation task (\cref{fig:sim-compare}c). Here, \simName's accurate contact force solving enables faithful reproduction of gel pad-object handle interactions, closely matching real-world execution patterns (detailed analysis in \cref{sec:rob:manip}).

Finally, the computational efficiency of \simName emerges from our optimized implementation of the \ac{abd} and \ac{ipc} algorithms, enabling unprecedented scaling capabilities. In a peg insertion task, \simName achieves parallel simulation of over 4096 environments on a single GPU with 80GB VRAM---representing a 64-fold improvement over SAPIEN-IPC~\cite{chen2024general}. This opens new possibilities for large-scale robotics simulation and learning; see also the comprehensive analysis in \cref{sec:eval:scaling}.

\begin{figure}[t!]
    \centering
    \includegraphics[width=\linewidth]{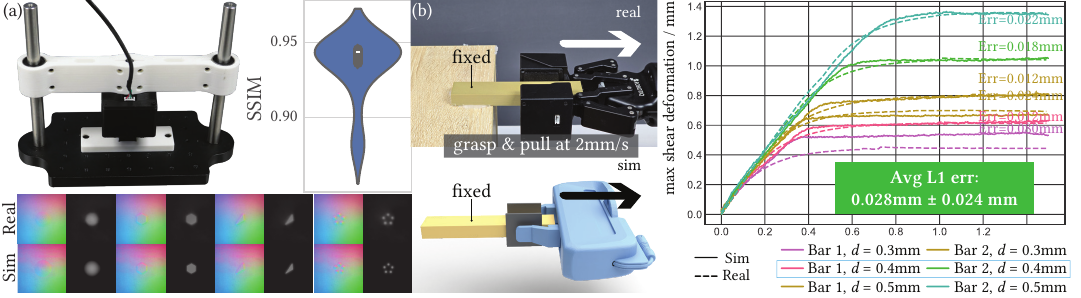}
    \caption{\textbf{Evaluations on the simulation precision.} (a) Data collection setup, examples of the real and simulated tactile patterns, and the sim-real SSIM distribution (violin plot) in the tactile signal evaluation. (b) Data collection setup and the shear deformation magnitude trajectories in the frictions simulation evaluation.}
    \label{fig:precision-eval}
\end{figure}

\subsection{Tactile Signal Simulation}\label{sec:eval:tactile-signal}

We evaluate the fidelity of tactile signals generated by \simName with real-world samples. We press a calibrated GelSight-type sensor perpendicullarly on 18 objects from a standard tactile shape testing dataset~\cite{gomes2021generation} on a real-world setup (\cref{fig:precision-eval}(a)) and in \simName; see \cref{supp:signal-test} for setup details.

The qualitative and quantitative comparisons shown in \cref{fig:precision-eval}(a) demonstrate \simName's ability to produce highly realistic tactile patterns, achieving an average SSIM of 0.93 across all test objects. Minor variations between simulated and real signals primarily stem from manufacturing tolerances in the 3D-printed objects and challenges in precise camera calibration. Despite these practical limitations, the results establish \simName's capability to generate high-fidelity tactile signals suitable for \acp{vbts}, with simulated patterns closely matching experimental measurements.

\subsection{Precision on Frictions and Shear Deformation}\label{sec:eval:friction}

We further investigate \simName's precision on solving frictions and shear deformation. \cref{fig:precision-eval}(b) shows the evaluation setup: A \acs{vbts}-gripper grasps a fixed bar with various forces (causes tactile depth $d$) then pulls back at $\SI{2}{mm/s}$, recording gel deformation represented by the marker deformations. We calibrate the object's friction coefficient with one record and use two others with various $d$s to compare the errors between simulated and real-world trajectories. Small sim-real errors (avg $\SI{28}{\mu m}$) tested on two bars with various frictions highlights \simName's fidelity on gel deformation and frictions in both static and slipping cases.

\subsection{Multi-environment Simulation}\label{sec:eval:scaling}

Efficient and stable parallel simulation of multiple environments is crucial for scaling up synthetic data collection across diverse downstream tasks. To evaluate \simName's capabilities in this regard, we designed three test cases of increasing complexity: (i) a dual-sensor (139 nodes per gel pad) peg-insertion task adapted from SAPIEN-\ac{ipc}~\cite{chen2024general}, (ii) the peg-insertion task with higher solution (1.5k nodes per gel pad), and (iii) a grasping task using a customized five-fingered dexterous hand equipped with 17 gel pads wrapped around the hand links (5k nodes). These tasks involve continuous contact and gel pad deformation, shedding light on how the simulator would perform on various robotic tasks. \cref{supp:parallel-test} provide more details on these tasks.

\begin{figure}[t!]
    \centering
    \centering
    \includegraphics[width=\linewidth]{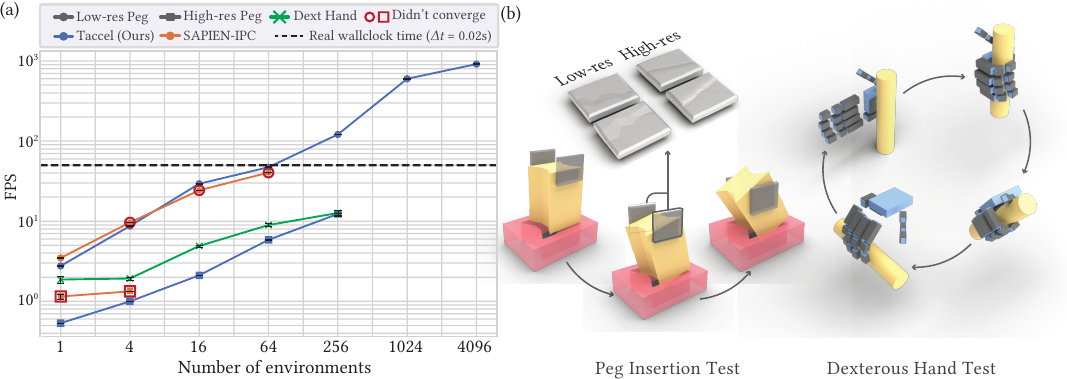}
    \caption{\textbf{Parallel simulation performance analysis across environment scaling.} (a) FPS achieved by \simName (FP64) and SAPIEN-\ac{ipc} (FP32) on an NVIDIA H100 80G GPU. (b) Task visualizations.}
    \label{fig:perf}
\end{figure}

With a single NVIDIA H100 80G GPU, we benchmarked \simName against SAPIEN-\ac{ipc}. The results in \cref{fig:perf} showcase \simName's superior performance: in the low-resolution peg-insertion task, \simName achieves 915 FPS (18.30$\times$ wallclock time) while managing over 4096 parallel environments---a 16-fold improvement over the baseline using the same GPU memory, enabled by our efficient parallelization implementation. The high-resolution test further demonstrates \simName's advantages, maintaining both stability and precision while consistently outperforming the baseline. Even in the complex dexterous hand scenario, \simName efficiently handles full-hand tactile sensing, achieving 12.67 FPS across 256 environments. We observed that SAPIEN-\ac{ipc}'s use of FP32 precision leads to convergence issues when solving contact forces in \cref{eqn:fullspace_ipc}, particularly in the logarithmic barrier energy term calculations, as indicated by red outlines in \cref{fig:perf}.

\section{Applications in Tactile Robotics}\label{sec:robotics}

We demonstrate \simName's capabilities across three tactile-informed robotic tasks: training object classification models with synthetic data (\cref{sec:rob:cognition}), generating a large-scale dataset through parallel grasp simulations (\cref{sec:rob:grasp}), and manipulating articulated objects (\cref{sec:rob:manip}). These applications showcase how \simName enables precise robotic simulation and scalable synthetic data generation.

\begin{figure}[t!]
    \centering
    \includegraphics[width=\linewidth]{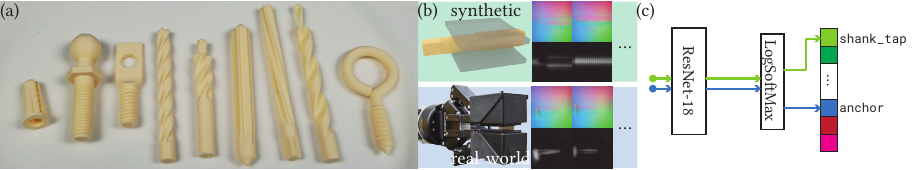}
    \caption{\textbf{Object classification sim-to-real pipeline.} (a): 10 mechanical parts used with diverse geometries. (b) Simulation and real-world data collection setup. (c) Classification network architecture.}
    \label{fig:results:bolt-cls}
\end{figure}

\subsection{Learning Object Classification Models}\label{sec:rob:cognition}

To demonstrate \simName's ability to generate synthetic training data that generalizes to real-world scenarios, we developed a tactile-based object classification system. As shown in \cref{fig:results:bolt-cls}, our approach trains a \ac{dnn} to classify objects using high-resolution tactile signals.

Following Yang \etal~\cite{yang2023hand}, we selected 10 mechanical parts with distinct fine-grained geometries (illustrated in \cref{fig:results:bolt-cls}, top). We collect a training dataset 4K tactile depth maps of a tactile sensor pressing on them within \simName, each with object pose and tactile depth randomly sampled. We also collect a real-world test set that consists of 160 depth maps for testing. \cref{supp:class-setup} illustrates more details for the data collection protocol. We trained a ResNet-18 model~\cite{he2016deep} for 10-category object classification using the simulated depth images. To enhance robustness, we augmented the depth maps through random affine transformations, morphological operations (erosion and dilation), and Gaussian filtering. The model is then directly evaluated them on the real-world samples, with each sample tested 4 times with random shear deformation. As shown in \cref{tab:obj-perception-results}, our model achieved 86.50\% accuracy on the synthetic test set and 70.94\% on real-world samples without any domain adaptation. This modest sim-to-real gap demonstrates \simName's capability to generate precise tactile signals that enable data-efficient training of transferable tactile perception models.

\subsection{Robotic Grasping with Tactile Sensors}\label{sec:rob:grasp}

\begin{figure}[b!]
    \centering
    \includegraphics[width=0.7\linewidth]{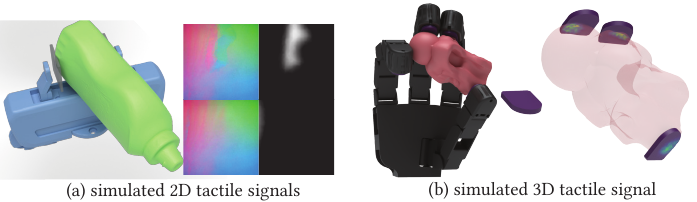}
    \caption{\textbf{Examples of the synthesized grasps and the simulated tactile signals.}}
    \label{fig:results:dfc-grasp-sim}
\end{figure}

We investigate robotic grasping across different hand configurations with varying tactile sensor arrangements. We first extend the DFC algorithm~\cite{liu2021synthesizing} for generating contact-oriented grasping poses for four robotic hands, then simulate their tactile responses within \simName, as illustrated in \cref{fig:results:dfc-grasp-sim}. We generate grasps on 10 diverse objects from ContactDB~\cite{brahmbhatt2019contactdb}, YCB~\cite{calli2015ycb}, and adversarial object~\cite{mahler2017dex} datasets. For each object, we generated grasps using 4 different robotic hands, producing \~14k total grasps. These were simulated in \simName to generate tactile signals, with key metrics summarized in \cref{tab:obj-perception-results}. \cref{supp:results:dfc-grasp-sim} provides an additional visualizations for the grasps across 4 robotic hands. \cref{supp:grasp} explains the details of the modification to DFC.

By simulating robotic grasps with tactile sensors, we can obtain synthetic data extending the existing robotic grasping datasets with tactile perception capabilities, serving as a foundation for various robotic tasks. To demonstrate its utility, we implemented the object classification task described in \cref{sec:rob:cognition}, adapting it to use the deformed coat's point cloud as input and PointNet as the feature extractor.

The classification performances in \cref{tab:obj-perception-results} reveals an interesting trade-off: while robots differ in sensor count and sensing area, higher dexterity (Allegro Hand) enables better object contact despite fewer sensors. This enhanced contact leads to superior classification accuracy, highlighting the balance between sensor count and dexterity in tactile hand design. These findings demonstrate \simName's value in validating robotic hand designs before physical fabrication.

\begin{table}[t!]
    \centering
    \caption{\textbf{Object classification and sim-to-real performance.} Performance metrics include number of tactile sensors ($N$), \acf{dof}, total sensing area (S.A.), average sensor-object contact area (C.A.) with percentage of total sensing area in parentheses, and classification accuracy (Acc.).}
    \begin{subfigure}{0.49\linewidth}
    \label{tab:obj-perception-results}
    \centering
    \caption{Sim-to-real results on mechanical part recognition.}
    \resizebox{\linewidth}{!}{
    \begin{tabular}{cccccc}
        \toprule
        \textbf{Data} & \textbf{\ac{dof}} &\textbf{$N$} & \textbf{S.A.} / $\SI{}{cm}^2$ & \textbf{C.A.} / $\SI{}{cm}^2$ & \textbf{Acc.} $\uparrow$ \\
        \midrule
        Mech. & 2 & 2 & 32.0 & 4.77 (14.92\%) & 86.50\% \\
        Mech. (Real) & 2 & 2 & 32.0 & 5.69 (17.78\%) & 70.94\% \\
        \bottomrule
    \end{tabular}
    }
    \end{subfigure}
    \begin{subfigure}{0.49\linewidth}
    \label{tab:obj-perception-results-hands}
    \centering
    \caption{Results on various robotic hand configurations.}
    \resizebox{\linewidth}{!}{
    \begin{tabular}{cccccc}
        \toprule
        \textbf{Data} & \textbf{\ac{dof}} &\textbf{$N$} & \textbf{S.A.} / $\SI{}{cm}^2$ & \textbf{C.A.} / $\SI{}{cm}^2$ & \textbf{Acc.} $\uparrow$ \\
        \midrule
        Gripper & 2 & 2 & 32.0 & 5.05 (15.78\%) & 44.56\% \\
        Robotiq-3F & 8 & 3 & 27.0 & 4.98 (18.43\%) & 44.61\% \\
        Allegro & 16 & 4 & 23.0 & \textbf{7.67 (33.34\%)} & \textbf{54.30\%} \\
        F-TAC~\cite{zhao2024embedding} & 15 & 17 & 59.7 & 4.00 (6.700\%) & 42.54\% \\
        \bottomrule
    \end{tabular}%
    }
    \end{subfigure}
\end{table}

\subsection{Articulated Object Manipulation}\label{sec:rob:manip}

We consider articulated object manipulation, a challenging task where tactile perception provides critical feedback on hand-object contact, informing object articulation and guiding robot actions~\cite{jara2014control,zhao2024tac}. \cref{supp:tacman} detailedly explain the algorithm. 

\begin{figure}[b!]
    \centering
    \includegraphics[width=\linewidth]{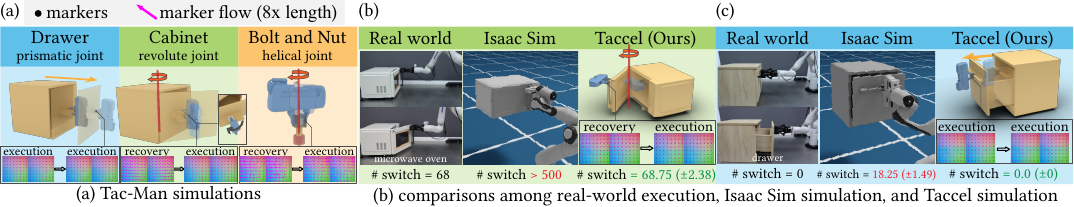}
    \caption{\textbf{Tac-Man manipulation simulation.} (a) Demonstration of Tac-Man's execution-recovery cycles on three articulated objects. (b-c) Three-way comparison among real-world execution, Isaac Sim implementation from Zhao \etal~\cite{zhao2024tac}, and \simName. The execution-recovery switch counts (\#switch) demonstrate \simName's accuracy in replicating real-world behavior ($1.10\%$ error), compared to Isaac Sim's higher counts.}
    \label{fig:tac-man}
\end{figure}

Tac-Man's effectiveness relies heavily on gel pad deformation for motion adaptation and tactile feedback. While the original implementation by Zhao \etal used rigid body simulation with gripper compliance approximations, it couldn't authentically replicate gel pad deformation, contact dynamics, and tactile feedback, resulting in significant sim-to-real gaps during large-scale verification. \simName overcomes these limitations through accurate simulation of sensor-object contact and gel pad deformation. We simulate Tac-Man on three types of articulated objects: drawers with prismatic joints, cabinets with revolute joints, and bolt-nut pairs with helical joints. \cref{fig:tac-man}(a) shows the manipulation sequences and corresponding tactile signals.

To evaluate sim-real correspondence, we compared our simulation against real-world Tac-Man execution using a microwave (revolute joint) and drawer (prismatic joint), shown in \cref{fig:tac-man}(b-c). For comprehensive comparison, we also tested Tac-Man's official implementation on these objects. As demonstrated in \cref{fig:tac-man}, \simName faithfully reproduces real-world execution patterns, with execution-recovery switch counts averaging 68.75 and 0.0 for revolute and prismatic settings, respectively---remarkably matching real-world observations. While Isaac Sim successfully simulated the manipulation, its dynamics gap resulted in substantially higher switch counts. These results highlight \simName's capability to authentically replicate physical interactions in manipulation scenarios.

\section{Summary}\label{sec:summary}

We present \simName, a flexible and high-performance simulator for \ac{vbts}-integrated robots. Its user-friendly APIs enables precise and efficient simulation of complex tactile robotic tasks with realistic tactile signals. \simName excels in capturing intricate deformation and contact dynamics of soft gel pads with unprecedented stability, while supporting thousands of parallelized environments. These capabilities also position \simName as a powerful tool for validating hand designs and sensor configurations before fabrication, potentially reducing development time and costs.

\paragraph*{Limitations}

The computational demands of large-scale simulation remain challenging for \simName. A major bottleneck is the PCG-based linear system solving that we will further optimize. Additional optimizations include carefully relaxing convergence tolerances, simplifying simulation protocols, and using larger timesteps, enabled by IPC’s unconditionally stable solver.

\paragraph*{Broader Impact} This paper focuses on fundamental research, with limited direct societal impact.

\begin{ack}
    This work is supported in part by the National Science and Technology Major Project (2022ZD0114900), the National Natural Science Foundation of China (62376009), the Beijing Nova Program, the State Key Lab of General AI at Peking University, the PKU-BingJi Joint Laboratory for Artificial Intelligence, and the National Comprehensive Experimental Base for Governance of Intelligent Society, Wuhan East Lake High-Tech Development Zone. We thank Lei Yan (LeapZenith AI Research) for his support in mechanical engineering, Zhenghao Qi (PKU, THU), Sirui Xie (PKU) for their assistance in setting up experiments, and Ms. Hailu Yang (PKU) for her help in procuring raw materials.
\end{ack}

\balance
\bibliographystyle{plain}
\bibliography{reference_header,reference}

\clearpage
\appendix
\renewcommand\thefigure{A\arabic{figure}}
\setcounter{figure}{0}
\renewcommand\thetable{A\arabic{table}}
\setcounter{table}{0}
\renewcommand\theequation{A\arabic{equation}}
\setcounter{equation}{0}
\pagenumbering{arabic}%
\renewcommand*{\thepage}{A\arabic{page}}
\setcounter{footnote}{0}

\section{Unified ABD-IPC in \simName}\label{supp:formulation}

In \simName, robot links are efficiently modeled as affine bodies to capture their primarily rigid motion, while \acp{vbts} are simulated as soft bodies to accurately represent their deformation mechanics. This natural division allows \simName to balance computational efficiency with physical accuracy while maintaining consistent contact handling through \ac{ipc}.

We provide a more detailed derivation of the unified ABD-IPC algorithm underlying \simName, as previously briefed in~\cref{sec:ipc}.

\subsection{Soft Body Dynamics}

Recall that we consider $n_s$ tetrahedralized soft bodies with $N_s$ vertices: $\vf{x}_1, \vf{x}_2, ..., \vf{x}_{N_s}$. The stacked position vector $\bx = [\vf{x}^T_1, \vf{x}^T_2, ..., \vf{x}^T_{N_s}]^T \in \mathbb{R}^{3N_s}$ represents the system state, and thus the system's Lagrangian is $L(\bx, \dot{\bx}) = T(\bx, \dot{\bx}) - V(\bx)$ with the kinetic energy $T(\bx, \dot{\bx}) = \frac{1}{2} \dot{\bx}^T \bM \dot{\bx}$, given the mass matrix $\bM \in \mathbb{R}^{3N_s\times3N_s}$. The potential energy $V(\bx)$ is composed of the elastic energy $\Phi(\bx)$ utilizing the Neo-Hookean constitutive model for hyperelastic materials (characterized by Young's modulus $\calE$ and Poisson's ratio $\nu$), and the external forces $E_{\text{ext}}(\bx)$. The elastic energy is defined as $\Phi(\bx)=\int_{\Omega} \Psi(\bx)d\bx$, where $\Psi(\bx)$ denotes elastic energy density over the volume region $\Omega$ of all objects in rest configuration.

\subsection{Time Stepping}\label{supp:sim:time-stepping}

Substituting $L(\bx, \dot{\bx})$ into the Euler-Lagrange equation $\frac{\partial L}{\partial \bx}(\bx, \dot{\bx}) - \frac{d}{dt}\frac{\partial L}{\partial \dot{\bx}}(\bx, \dot{\bx}) = 0$ yields the governing dynamics:
\begin{equation}
    \bM\ddot{\bx} = -\frac{dV}{d\bx}(\bx).
    \label{eqn:euler-lagrangian}
\end{equation}

We temporally discretize \cref{eqn:euler-lagrangian} using backward Euler:
\begin{equation}
    \frac{\bx^{n+1} - \bx^{n}}{\Dt} = \dot{\bx}^{n + 1}, \quad \frac{\bM(\dot{\bx}^{n+1} - \dot{\bx}^{n})}{\Dt} = -\frac{dV}{d\bx}(\bx^{n+1}),
\end{equation}
where time is discretized into steps $\{t_n = n\Dt: n\in \mathbb{N}\}$ with step size $\Dt > 0$, and $\bx^{n}=\bx(t_n)$. Under this discretization, \cref{eqn:euler-lagrangian} can be formulated as:
\begin{equation}
    \frac{d}{d \bx}\left(E_{\text{IP}}(\bx^n)\right) = 0.
\end{equation}

If we define the incremental potential energy of the constrained system as:
\begin{equation}
    \begin{aligned}
        E_{\text{IP}}(\bx) &= \frac{1}{2}(\bx - \bx^{n} - \Dt\dot{\bx}^{n})^T\bM(\bx - \bx^{n} - \Dt\dot{\bx}^{n})\\
        &+ \Dt^2V(\bx),
    \end{aligned}
\end{equation}
then the general simulation problem in a conservative system can be reformulated as the minimization problem:
\begin{equation}
    \bx^{n+1} = \argmin_{\bx} E_{\text{IP}}(\bx).
\end{equation}

\subsection{Frictional Contact}

We employ \ac{ipc}~\cite{li2020incremental} to handle contact interactions. The method operates on surface contact pairs $\mathcal{B}$, comprising point-triangle and edge-edge pairs from the surface meshes of soft and affine objects. For each contact pair $k\in \mathcal{B}$ with distance $d_k>0$, \ac{ipc} defines two key energy terms. First, a barrier energy that prevents interpenetration:
\begin{equation}
    b(d_k(\bx))=-\left(d_k - \hat{d}\right)^2\log(\frac{d_k}{\hat{d}})I_{\{d_k \in (0, \hat{d})\}}(d_k),
    \label{eqn:ipc_energy}
\end{equation}
where $\hat{d}>0$ is the distance threshold for contact force activation and $I(\cdot)$ is the indicator function.

Second, an approximated friction potential energy:
\begin{equation}
    D_k(\bx, \bx^{n}) = \mu\lambda_k^{n}f_0(\lVert \vf{u}_k\rVert),
\end{equation}
where $\bx^{n}$ represents the configuration at the previous timestep $t_n$, $\lambda_k^{n}$ is the magnitude of the lagged normal contact force, and $\vf{u}_k\in \mathbb{R}^2$ denotes the tangential relative displacement in the local contact frame.

The friction transition function $f_0(x)=\int_{\epsilon_v\Dt}^x f_1(y)dy + \epsilon_{v}\Dt$ uses:
\begin{equation}
    f_1(y) = 
    \begin{cases}
        -\frac{y^2}{\epsilon_v^2\Dt^2} + \frac{2y}{\epsilon_v\Dt}, &y\in(0, \Dt\epsilon_v),\\
        1, & y\geq \Dt\epsilon_v,
    \end{cases}
\end{equation}
where $\epsilon_v > 0$ serves as a velocity threshold distinguishing between static and dynamic friction regimes.

These contact and friction terms augment our incremental potential energy:
\begin{equation}
    E_{\text{\ac{ipc}}}(\bx) = E_{\text{IP}}(\bx) + \Dt^2B(\bx) + \Dt^2D(\bx, \bx^{n}),
    \label{suppeq:fullspace_ipc}
\end{equation}
with {\small$B(\bx)=\kappa \sum_{k\in \mathcal{B}} A_k b(d_k(\bx))$, $D(\bx, \bx^n)=\sum_{k\in\mathcal{B}} D_k(\bx, \bx^n)$}, where $\kappa > 0$ controls contact stiffness.

\subsection{\texorpdfstring{\acs{abd}}{} and Unified Simulation}\label{supp:affine}

For $n_a$ affine bodies, we introduce a reduced coordinate space $\by \in \mathbb{R}^{12 n_a}$ with an embedding map $\phi: \mathbb{R}^{12 n_a} \rightarrow \mathbb{R}^{3N_a}$ that projects reduced coordinates to full-space vertices $\phi(\mf{\by})$~\cite{lan2022affine}, where $N_a$ denotes the total vertex count of affine bodies' surface meshes. Each affine body uses 12 \ac{dof}: three for translation ($\mathbb{R}^{3}$) and nine for affine deformation ($\mathbb{R}^{3\times 3}$).
\begin{equation}
    \begin{aligned}
        & T(\by, \dot{\by}) = \frac{1}{2} \dot{\bx}^T \bM \dot{\bx} = \frac{1}{2} \dot{\phi}(\by)^T \bM\dot{\phi}(\by) \\&= \frac{1}{2} (\bJ\dot{\by})^T \bM (\bJ\dot{\by}) = \frac{1}{2}\dot{\by}^T(\bJ^T\bM\bJ)\dot{\by}= \frac{1}{2}\dot{\by}^T \mf{M^y} \dot{\by},
    \end{aligned}
\end{equation}
where $\bJ = \frac{\partial \phi}{\partial \by}\in \mathbb{R}^{3N_a \times 12 n_a }$ is the Jacobian, $\bM$ is the full-space mass matrix, and $\mf{M^y} = \bJ^T\bM\bJ$ is the reduced-space mass matrix. The potential energy $V(\by)$ includes an As-Rigid-As-Possible (ARAP) term $\Phi^{\by}(\by)=\Phi^{\bx}(\phi(\bx))$ with high stiffness $\kappa_s$ to limit deformation, plus external forces $E_{\text{ext}}(\by)$.

Combining with \cref{suppeq:fullspace_ipc}, we obtain the unified affine-deformable coupled \ac{ipc} energy~\cite{chen2022unified} for the full system state $\{\by; \bx \}\in \mathbb{R}^{12 n_a + 3 N_s}$:
\begin{equation}
    \begin{aligned}
        E_{\text{\ac{ipc}}}(\by; \bx) &= E_{\text{IP}}(\bx) + E_{\text{IP}}(\by)+ \Dt^2B(\phi(\by); \bx) \\
        &+ \Dt^2D(\phi(\by); \bx, \phi(\by^{n}); \bx^{n}),
    \end{aligned}
    \label{suppeq:unified_ipc}
\end{equation}
where $E_{\text{IP}}(\by)$ is defined as:
\begin{equation}
    \begin{aligned}
        E_{\text{IP}}(\by) &= \frac{1}{2}(\by - \by^{n} - \Dt\dot{\by}^{n})^T\mf{M^y}(\by - \by^{n} - \Dt\dot{\by}^{n})\\
        &+ \Dt^2V(\by).
    \end{aligned}
\end{equation}

The next timestep's configuration follows from minimizing this barrier-augmented incremental potential:
\begin{equation}
    \by^{n+1};\bx^{n+1} = \argmin_{\by;\bx} E_{\text{\ac{ipc}}}(\by;\bx).
    \label{suppeq:unified_ipc_variational}
\end{equation}

\subsection{Kinematic Constraints}\label{supp:sim:kine-constraint}

The kinematic constraints is expressed as
\begin{equation}
    \bS^{\bx} \bx = \bs^{\bx}, \bS^{\by} \by = \bs^{\by},
\end{equation}
where $\bS^{\bx} \in \mathbb{R}^{c^{\bx} \times 3 N_s}, \bs^{\bx} \in \mathbb{R}^{3 N_s}$ for soft bodies, and $\bS^{\by} \in \mathbb{R}^{c^{\by} \times 12 n_a}, \bs^{\by} \in \mathbb{R}^{12 n_a}$ for affine bodies. The constraints are applied by selecting the constrained \acp{dof} of the state vectors and specifying the constraint values with. To enforce these constraints, we employ the Augmented Lagrangian method by augmenting $E_{\text{\ac{ipc}}}$ to:
\begin{equation}
    \begin{aligned}
        & E_{\text{\ac{ipc}}}^{\text{AL}}(\by; \bx) = E_{\text{\ac{ipc}}}(\by; \bx) \\ 
        &+  \Vert (\bS^{\bx} \bx - \bs^{\bx} )^T\mf{\lambda}^{\bx}\Vert_2^2 + \Vert (\bS^{\by} \by - \bs^{\by} )^T\mf{\lambda}^{\by}\Vert_2^2,
    \end{aligned}
    \label{suppeq:unified_ipc_AL}
\end{equation}
where $\mf{\lambda}^{\bx} \in \mathbb{R}^{c^{\bx}}$ and $\mf{\lambda}^{\by} \in \mathbb{R}^{c^{\by}}$ are Lagrangian multipliers.

Optimizing $E_{\text{\ac{ipc}}}^{\text{AL}}(\by; \bx)$ yields the solution to the constrained system:
\begin{align}
    \by^{n+1};\bx^{n+1} &= \argmin_{\by;\bx} E_{\text{\ac{ipc}}}(\by;\bx), \label{eq:abdipc-kine-constraint} \\
    \text{s.t.}  \quad \bS^{\bx} \bx &= \bs^{\bx} \quad{}\text{and}\quad{} \bS^{\by} \by = \bs^{\by}. \label{suppeq:abdipc-kine-constraint-st}
\end{align}

\section{Additional Related Work}\label{supp:related}

\subsection{Robot Tactile Sensors}

Tactile sensing plays a fundamental role in precise manipulation, as established by neuroscientific studies~\cite{westling1984factors,johansson2009coding,jiang2024capturing,billard2019trends}. This understanding has driven the development of artificial tactile sensing systems for robots~\cite{qu2023recent}. Among these, \acp{vbts} have gained prominence by offering high-resolution sensing with cost-effectiveness and operational simplicity~\cite{yuan2017gelsight,ward2018tactip,lin20239dtact,li2024minitac}. While these sensors have advanced robotic manipulation~\cite{she2021cable,lloyd2024pose,zhao2024tac}, their development remains constrained by the reliance on physical hardware experimentation. \simName addresses this limitation by providing a comprehensive simulation platform to accelerate research and development in tactile robotics.

\subsection{Tactile-Informed Robotic Tasks}

Tactile sensing enhances robotic capabilities across three fundamental domains through precise contact interaction measurements:

\paragraph*{Perception}

Tactile feedback enables sophisticated object understanding through contact-based sensing. Applications include shear and slip detection~\cite{yuan2015measurement,dong2017improved}, object classification and pose estimation~\cite{li2014localization,yang2023hand,suresh2024neuralfeels,bauza2023tac2pose}, material property inference~\cite{guo2023estimating,huang2022understanding}, and interaction reconstruction~\cite{suresh2024neuralfeels,yu2024dynamic,xu2023visual}. These perceptual capabilities form the foundation for advanced manipulation algorithms.

\paragraph*{Grasping}

Stable grasping requires precise control of contact forces to balance external loads~\cite{ferrari1992planning,siciliano2016robotics}. Tactile sensing provides direct force-torque feedback essential for diverse grasping strategies~\cite{liu2021synthesizing,li2024grasp,yao2023exploiting}. This tactile information complements vision-based approaches by enabling fine-grained contact monitoring and in-hand adjustments~\cite{calandra2018more,bronars2024texterity}.

\paragraph*{Manipulation}

Tactile feedback enables complex manipulation beyond basic pick-and-place operations. Applications include precision tasks like peg insertion~\cite{chen2024general}, object pivoting~\cite{hogan2020tactile}, and articulated object manipulation~\cite{zhu2020dark,billard2019trends}. Systems such as Tac-Man~\cite{zhao2024tac} and DoorBot~\cite{wangdoorbot} demonstrate how tactile sensing guides contact geometry understanding and articulation control. This sensing modality is particularly crucial for high-frequency object tracking during dexterous manipulation.

We validate \simName's capabilities through three representative applications: (i) multi-platform robotic grasping with both rigid and soft objects, (ii) object classification using purely synthetic training data with strong real-world transfer, and (iii) articulated object manipulation including drawers, cabinets, and bolt-nut assembly tasks, extending the Tac-Man framework~\cite{zhao2024tac}.

\section{Additional Details of Experiments}

\subsection{Real-world Data Collection for Tactile Signal Evaluation}\label{supp:signal-test}

For real-world data collection, our experimental setup consisted of a calibrated GelSight-type sensor, mounted on a vertical rack for precise movement control, as shown in \cref{fig:precision-eval}(a). A white mount is fixed in the center of the platform to hold the test objects.

We 3D-print the 18 objects from a standard tactile shape testing dataset~\cite{gomes2021generation} at 0.2mm layer height. For each object, we first fix it on the mount, slide the sensor along the rack to press its gel pad on the object, and put a \SI{500}{g} weight on the sensor to ensure an appropriate amount of gel deformation. We record the RGB tactile patterns five times and compute their mean image to partially remove the noise. We then replicated this pressing sequence in \simName using a high-resolution soft body gel pad (maximal cell volume $V_{\max} \approx 10^{-12} \text{m}^3$) to ensure signal fidelity. \cref{sec:eval:tactile-signal} compares the similarity between the real-world and simulated tactile signals.

\subsection{Multi-environment Simulation Test}\label{supp:parallel-test}

Our multi-environment simulation test involves three scripted tasks. First, we implemented a peg-insertion task adapted from SAPIEN-\ac{ipc}~\cite{chen2024general}, where two gel pads (139 nodes and 317 cells each) follow a scripted trajectory. The trajectory involves 200 simulation steps, where the sensors squeeze the peg and manipulate it around the hole. Next, we scaled this task to a higher resolution (1,533 nodes and 5,360 cells each), which results in a larger system to solve. Finally, to demonstrate \simName's potential for advanced robotics research, we created a scripted grasping task with a customized five-fingered dexterous hand equipped with 17 gel pads covering the entire hand, totaling 5,157 nodes and 14,311 cells. The trajectory also involves 200 simulation steps, with the hand approaches, grasps, lifts, maneuvers, and releases a stiff cylinder.

\subsection{Data Collection and Training Details for Learning Object Classification}\label{supp:class-setup}

In the object classification sim-to-real experiment, we collect tactile signals of grasping 10 3D-printed mechanical parts in both simulation and the real world. We also train an object classification model with the simulated data and evaluate it on the real data.

For each object, we simulated 200 grasp trials using a parallel gripper with randomized grasping poses. The gripper closes towards the object until the depth deformation exceeds a threshold randomly sampled from $\tau_d \sim U[0.5, 1.5] \SI{}{mm}$, and with each grasp yields 2 depth map samples after grasping. To ensure the granularity of the depth map, high-resolution gel pad models $\max V \geq 10^{-12}\SI{}{m^3}$. The depth maps $d_{(u, v)}$ extracted from these simulated tactile signals yielded approximately 4,000 training samples. We split the samples into a training set (85\%) and a validation set (15\%). The former is used to train the classification model, supervised by the NLL Loss, using the Adam optimizer with a learning rate 1e-4 for 100 epochs. During training, we apply an exponential LR scheduler.

For real-world validation, we collected tactile signals using a RobotiQ-2F85 parallel gripper equipped with GelSight-type sensors. The test objects were 3D printed at 0.2mm layer height to maintain high geometric fidelity. We gathered 8 grasps (16 depth maps) per object.

\subsection{Dexterous Grasping}\label{supp:grasp}

Our key modification to the DFC algorithm promotes perpendicular contact between gel pads and object surfaces, optimizing for downstream tasks that rely on tactile perception.

Following Liu \etal~\cite{liu2021synthesizing}, we synthesize grasping poses in the robot's joint space $q \in \calQ$ relative to the object frame by minimizing a modified Gibbs energy:
\begin{equation}
    E(O, q, T) = E_\mathrm{DFC} + \lambda_\mathrm{contact} E_\mathrm{contact} (O, q, T).
\end{equation}
The force-closure term $E_\mathrm{DFC}$ maintains its original formulation~\cite{liu2021synthesizing}, with added constraints to ensure gel pad penetration depth $\epsilon$ (typically $\SI{0.5}{mm}$). We introduce a new contact term $E_\mathrm{contact}$ that aligns the normals of gel pad contacts $c_i \in \partial^+\calG_i$ with their corresponding object surface normals $o_i = \argmin_{o \in \partial O} \Vert o - c_i \Vert$:
\begin{equation}
    E_\mathrm{contact} (O, q, T) = 1 - \left< c^\perp_i, o^\perp_i \right>,
\end{equation}
where $(\cdot)^\perp$ represents the surface normals of the gel pad and object surfaces.

\begin{figure}[t!]
    \centering
    \includegraphics[width=\linewidth]{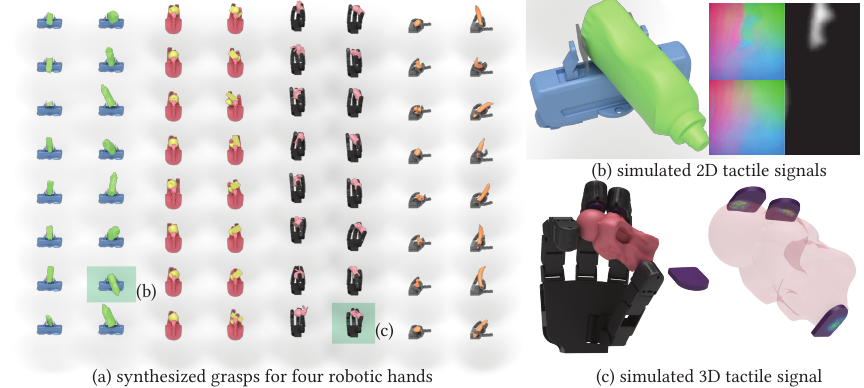}
    \caption{\textbf{Examples of the synthesized grasps and the simulated tactile signals in 2D and 3D.} (a) Diverse grasps are synthesized and simulated for different objects and robotic hands. (b) An example of the 2D tactile signals generated for a grasp with the Panda Hand. (c) An example of the 3D tactile signals generated for a grasp with the Allegro Hand with Digit sensors.}
    \label{supp:results:dfc-grasp-sim}
\end{figure}

\cref{supp:results:dfc-grasp-sim} shows the examples of the synthesized grasps and the simulated tactile signals in both 2D and 3D.

\subsection{Articulated Object Manipulation with Tac-Man}\label{supp:tacman}

In the Tac-Man framework~\cite{zhao2024tac}, the system alternates between \textit{execution} and \textit{recovery} phases. During execution, the system performs coarse manipulation actions (\eg, pulling backward) to gradually move the articulated object part. When the actual motion deviates from intended trajectories due to articulation constraints, the gel pad deforms, creating contact deviation reflected in marker flow magnitudes. Once these flows exceed threshold $\delta_0$, the system enters recovery mode to restore stable contact by reducing deviation, before resuming execution. This execution-recovery cycle typically requires tens of iterations to finish manipulation.

For our sim-real comparison, we manually created URDF models for the microwave oven and the drawer to match real-world geometries and kinematics, maintaining identical initial grasping poses across simulation and physical setups. Following Tac-Man's implementation, we set $\delta_0 = \SI{0.4}{mm}$ and $\alpha = 0.6$.

\end{document}